\documentclass[10pt,twocolumn,letterpaper]{article}

\usepackage{booktabs}   % 提供 \toprule, \midrule, \bottomrule, \cmidrule
\usepackage{multirow}   % 提供 \multirow 实现单元格跨行

\usepackage[table]{xcolor} 
\usepackage{adjustbox}
\usepackage{amsmath} 
\usepackage{graphicx}
\usepackage{pifont}
\definecolor{opensrcHdr}{HTML}{C8DDFB}
\definecolor{opensrcRow}{HTML}{EAF0FB} 
\definecolor{propriHdr}{HTML}{E0D4F5} 
\definecolor{propriRow}{HTML}{F2EDF9} 
\definecolor{spatialHdr}{HTML}{C8D5F0}
\definecolor{spatialRow}{HTML}{E8EDF8} 

\newcommand{\modelicon}[1]{\includegraphics[height=1em]{#1}}

\usepackage[pagenumbers]{cvpr} % To force page numbers, e.g. for an arXiv version

\definecolor{cvprblue}{rgb}{0.21,0.49,0.74}
\usepackage[pagebackref,breaklinks,colorlinks,allcolors=cvprblue]{hyperref}

\def\paperID{*****} % *** Enter the Paper ID here
\def\confName{CVPR}
\def\confYear{2026}

\title{World2Mind: Cognition Toolkit for Allocentric Spatial Reasoning in \\Foundation Models}

\author{
Shouwei Ruan$^{1}$, Bin Wang$^{2}$, Zhenyu Wu$^{1}$, Qihui Zhu$^{1}$, Yuxiang Zhang$^{2}$, \\Hang Su$^{3}$, Yubin Wang$^{2}$\thanks{Corresponding author.} \\
$^{1}$Institute of Artificial Intelligence, Beihang University \\
$^{1}$ Huawei Noah's Ark Lab \\
$^{3}$ Dept. of Comp. Sci. and Tech., Institute for AI, Tsinghua-Bosch Joint ML Center, \\THBI Lab, BNRist Center, Tsinghua University}

\begin{document}
\maketitle
\begin{abstract}
Achieving robust spatial reasoning remains a fundamental challenge for current Multimodal Foundation Models (MFMs). Existing methods either overfit statistical shortcuts via 3D grounding data or remain confined to 2D visual perception, limiting both spatial reasoning accuracy and generalization in unseen scenarios. Inspired by the spatial cognitive mapping mechanisms of biological intelligence, we propose \textbf{World2Mind}, a training-free spatial intelligence toolkit. At its core, World2Mind leverages 3D reconstruction and instance segmentation models to construct structured spatial cognitive maps, \textbf{empowering MFMs to proactively acquire targeted spatial knowledge regarding interested landmarks and routes of interest.} To provide robust geometric-topological priors, World2Mind synthesizes an \textbf{Allocentric-Spatial Tree (AST)} that uses elliptical parameters to model the top-down layout of landmarks accurately. To mitigate the inherent inaccuracies of 3D reconstruction, we introduce a three-stage reasoning chain comprising tool invocation assessment, modality-decoupled cue collection, and geometry-semantics interwoven reasoning. Extensive experiments demonstrate that World2Mind boosts the performance of frontier models, such as GPT-5.2, by 5\%$\sim$18\%. Astonishingly, relying solely on the AST-structured text, purely text-only foundation models can perform complex 3D spatial reasoning, achieving performance approaching that of advanced multimodal models.
\end{abstract}    
\section{Introduction}
\label{sec:intro}

\begin{figure*}[t]
    \centering
    \includegraphics[width=0.91\textwidth]{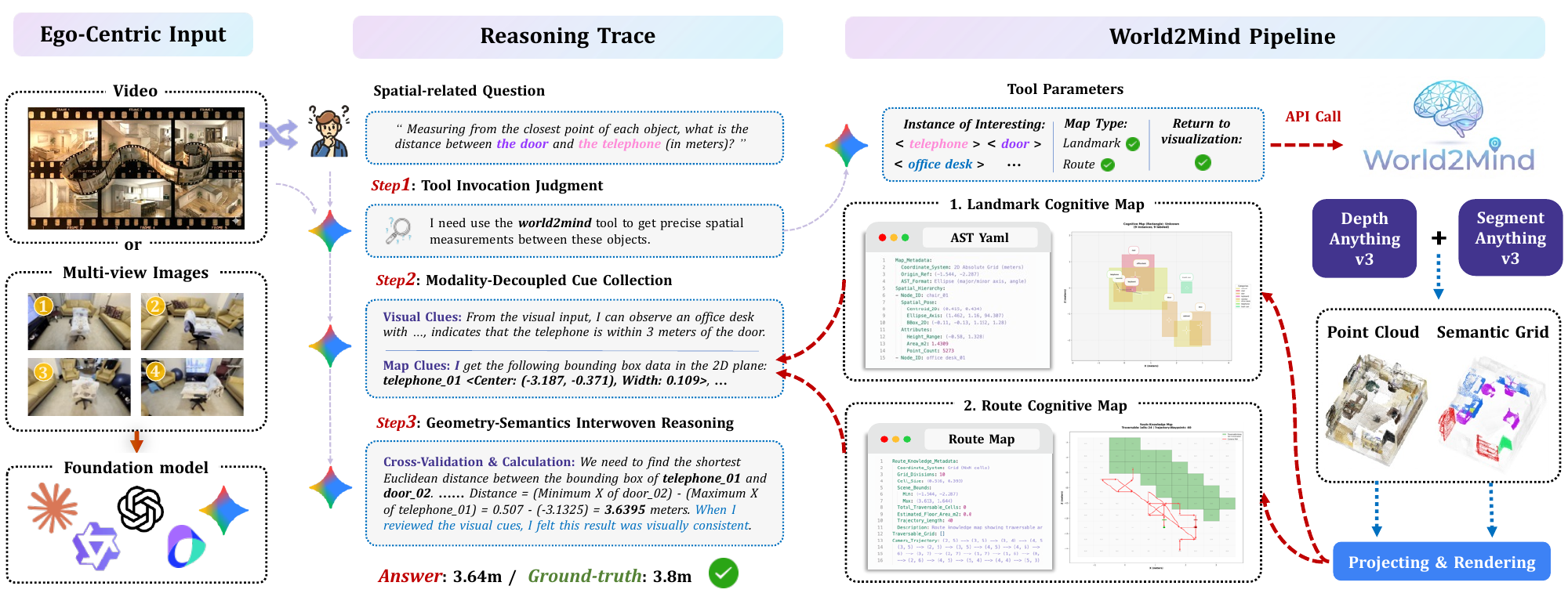}
    \caption{\textbf{Overview of foundation models performing allocentric spatial reasoning via the proposed World2Mind toolkit.} Given egocentric video or multi-view observations, the model first assesses the necessity of tool invocation and subsequently passes key parameters (\eg, instances of interest) to World2Mind to drive the generation of spatial cognitive maps. World2Mind integrates an efficient pipeline for 3D reconstruction and semantic-geometry alignment, returning the required structured spatial knowledge through targeted projection and rendering mechanisms. Furthermore, the model conducts geometry-semantics interwoven reasoning based on both the raw visual observations and the geometric cues provided by World2Mind, ultimately yielding highly reliable answers.}
    \label{fig:teaser}
    
\end{figure*}

% Multimodal Foundation Models (MFMs)~\cite{bai2025qwen3, openai2023gpt4v_systemcard, singh2025openai, gemini3pro} have recently demonstrated remarkable proficiency in general visual understanding and cross-modal reasoning~\cite{su2025thinking}. However, their performance often degrades significantly when deployed in embodied AI or complex spatial reasoning scenarios that require deep physical interaction~\cite{yang2025thinking, liu2024mindcube, lin2025mmsi, huang2025si, ramakrishnan2024does}. The fundamental root of this deficiency is their over-reliance on egocentric observations. Lacking the capacity to abstract and represent global spatial topology~\cite{yang2025thinking, liu2024mindcube}, these models face an insurmountable ``semantic-geometry gap'' when tackling tasks like distance estimation, viewpoint transformation, and path planning.

Although multimodal foundation models (MFMs)~\cite{bai2025qwen3, openai2023gpt4v_systemcard, singh2025openai, gemini3pro} excel in general visual understanding and cross-modal reasoning~\cite{su2025thinking}, they struggle significantly in embodied AI and complex spatial reasoning tasks requiring physical interaction~\cite{yang2025thinking, liu2024mindcube, lin2025mmsi, huang2025si, ramakrishnan2024does}. This deficiency stems from their over-reliance on egocentric observations and lacking the capacity to abstract global spatial topology~\cite{yang2025thinking, liu2024mindcube}, trapping MFMs in an insurmountable ``semantic-geometry gap'' in tasks like distance estimation, viewpoint transformation, and path planning.

% To improve MFMs' spatial reasoning capabilities, existing research predominantly follows two paradigms. \textbf{\textit{Training-based methods}}~\cite{spatialrgpt,spatialvlm,ma2025spatialreasoner} attempt to fine-tune models by synthesizing massive question-answering pairs on 3D-grounding datasets. However, this paradigm essentially forces the model to overfit statistical shortcuts~\cite{qi2025beyond, huang2025surprise3d} between visual information and geometric annotations. It fails to endow the model with genuine spatial cognition, leading to poor generalization to out-of-distribution scenarios~\cite{wu2025indoor}. An alternative approach introduces explicit 3D modalities~\cite{daxberger2025mm, chen2025sd, ning2025enhancing, wang2025n3d}, such as point clouds or depth information, for supervision. Yet, this exacerbates the challenges of multimodal encoding and inter-modal alignment~\cite{zhang2025point}. Recently, several \textbf{\textit{tool-based methods}}~\cite{zhang2026think3d, luo2026pyspatial, chen2025spacetools} have sought to incorporate large-scale 3D reconstruction and active novel-view rendering to gather reasoning cues. Nevertheless, these approaches are severely bottlenecked by rendering and reconstruction quality. More fundamentally, they remain tethered to low-level visual perception, failing to abstract convoluted geometric information into structured semantics accessible for high-level logical reasoning.

Current efforts to enhance MFMs' spatial reasoning primarily follow two paradigms. \textbf{\textit{Training-based methods}}~\cite{spatialrgpt,spatialvlm,ma2025spatialreasoner} fine-tune models on massive 3D-grounded QA pairs. However, this forces models to overfit statistical shortcuts~\cite{qi2025beyond, huang2025surprise3d} rather than acquiring genuine spatial cognition, leading to poor generalization in out-of-distribution scenarios~\cite{wu2025indoor}. Alternatively, introducing explicit 3D modalities~\cite{daxberger2025mm, chen2025sd, ning2025enhancing, wang2025n3d} exacerbates inter-modal alignment challenges~\cite{zhang2025point}. Meanwhile, recent \textbf{\textit{tool-based methods}}~\cite{zhang2026think3d, luo2026pyspatial, chen2025spacetools} rely on active rendering under large-scale 3D reconstruction. These are severely bottlenecked by reconstruction quality and remain tethered to low-level visual perception, failing to abstract geometric data into structured semantics for high-level logical reasoning.

% Diverging from current reasoning paradigms that rely solely on egocentric observations, Biological Intelligence (BI) provides an ideal blueprint: when reasoning and acting in the complex physical world, BI does not merely react passively to transient visual inputs. Instead, the biological brain intrinsically performs a coordinate transformation from an egocentric to an allocentric perspective~\cite{burgess2006spatial}, constructing a global \textbf{\textit{cognitive map}} that is entirely independent of the current viewpoint~\cite{o1978hippocampus, schiller2015memory}, which serves as the cornerstone for strategic mental simulation and advanced reasoning~\cite{bellmund2018navigating}. This mental simulation mechanism is underpinned by solid neuroscientific evidence: mammals, including humans, efficiently achieve this leap and application in spatial representation via place cells in the hippocampus~\cite{o1971hippocampus, eichenbaum2017role, broadbent2004spatial} and grid cells in the entorhinal cortex~\cite{hafting2005microstructure}.

Biological Intelligence (BI) offers an ideal blueprint to break the shackles of egocentric observation. Rather than reacting passively to transient visual inputs, the biological brain intrinsically transforms egocentric views into an allocentric perspective~\cite{burgess2006spatial}. Supported by place cells in the hippocampus~\cite{o1971hippocampus, eichenbaum2017role, broadbent2004spatial} and grid cells in the entorhinal cortex~\cite{hafting2005microstructure}, mammals construct a global \textbf{\textit{cognitive map}} entirely independent of their egocentric viewpoint~\cite{o1978hippocampus, schiller2015memory}. This map forms the cornerstone for strategic mental simulation and advanced reasoning~\cite{bellmund2018navigating}.

% To bridge the gap between foundation models and Biological Intelligence (BI) in spatial representation and reasoning, we propose \textbf{\textit{World2Mind}}, a plug-and-play spatial cignition toolkit that equips foundation model with human-like mental simulation and reasoning capabilities. Within World2Mind, we integrate an efficient geometry-semantics alignment pipeline. By leveraging pre-training Visual geometry models~\cite{wang2025vggt, lin2025depth, wang2025pi} and instance segmentation~\cite{carion2025sam} to extract semantic voxel grids of scenes, World2Mind constructs two core spatial mental representations: \textbf{\textit{1)}} Route Cognitive Map for predicting path passability, and \textbf{\textit{2)}} Landmark Cognitive Map for representing object topology. To enhance scalability, we encapsulate World2Mind into an efficient toolset. Specifically, models can dynamically interact with World2Mind by specifying customizable parameters (\eg, the list of instances of interest, the type of spatial knowledge required, and whether to retrieve map visualizations, \etc), thereby proactively acquiring spatial knowledge regarding specific landmarks and routes on demand.

To bridge this gap between foundation models and BI in spatial representation and reasoning, we propose \textbf{\textit{World2Mind}}, a \textbf{plug-and-play spatial cognition toolkit} that equips models with human-like mental simulation capabilities. World2Mind integrates an efficient geometry-semantics alignment pipeline, leveraging pre-trained visual geometry~\cite{wang2025vggt, lin2025depth, wang2025pi} and instance segmentation models~\cite{carion2025sam} to extract semantic voxel grids. From these, it constructs two core representations: \textbf{\textit{1)}} a Route Cognitive Map for passability prediction, and \textbf{\textit{2)}} a Landmark Cognitive Map for object topology. Encapsulated as an accessible toolset, World2Mind enables models to dynamically specify parameters (\eg, instances of interest, required spatial knowledge, and map visualizations) to proactively acquire targeted allocentric spatial knowledge on demand.

\begin{table*}[t]
\centering
\caption{\textbf{Main results on the VSI-Bench~\cite{yang2025thinking} benchmark (Tiny subset).}}
\label{table:visbench_tiny}
\renewcommand{\arraystretch}{1.0}
\setlength{\tabcolsep}{3.6pt}
\small
\begin{adjustbox}{width=\textwidth}
\begin{tabular}{@{}l ccccccccc@{}}
\toprule
\multirow{2}{*}{\textbf{Models}} & \multirow{2}{*}{\textbf{Avg.}} & \multicolumn{4}{c}{\textbf{Numerical Answer}~(\%)} & \multicolumn{4}{c}{\textbf{Multiple-Choice Answer}~(\%)}  \\
\cmidrule(lr){3-6} \cmidrule(lr){7-10}
 & & \textbf{\small Obj. Count} & \textbf{\small Abs. Dist.} & \textbf{\small Obj. Size} & \textbf{\small Room Size} & \textbf{\small Rel. Dist.} & \textbf{\small Rel. Dir.} & \textbf{\small Route Plan} & \textbf{\small Appr. Order} \\
\midrule
% ======================== Baseline ========================
\rowcolor{propriHdr!75}
\multicolumn{10}{c}{\textbf{\textit{Frontier models w/o. World2Mind}}} \\
% \modelicon{icons/qwen.png}~Qwen-3.5-397B & & &  &  &  &  &  &  &  \\
\modelicon{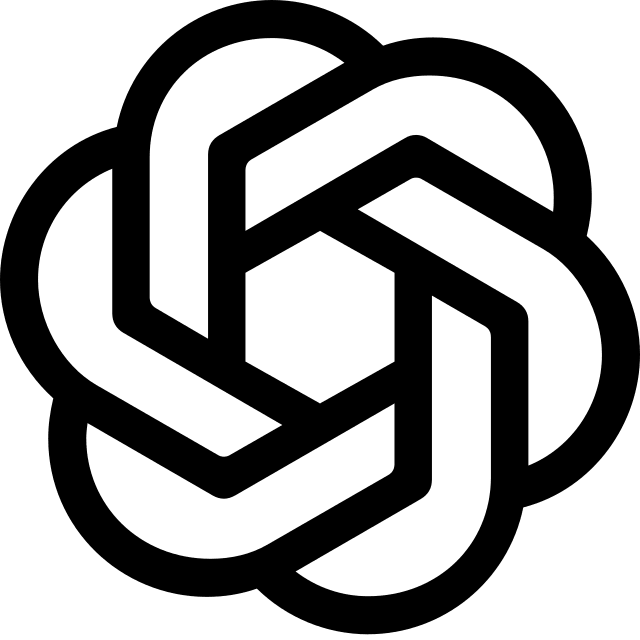}~GPT-5.2 & 46.7 & 52.5 & 34.9 & 67.5 & 50.6 & 42.0 & 40.7 & 34.7 & 51.0 \\
\modelicon{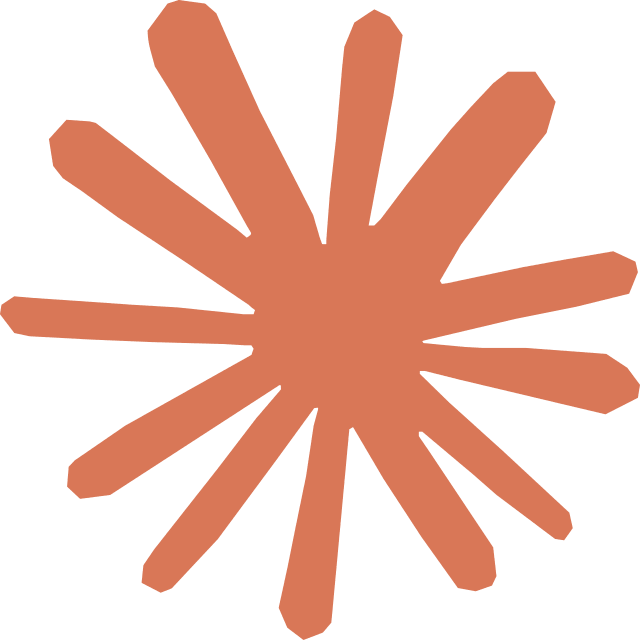}~Claude-4.6-Opus & 38.4 & 46.9 & 18.5 & 62.1 & 26.8 & 40.0 & 47.2 & 34.7 & 30.6 \\
\modelicon{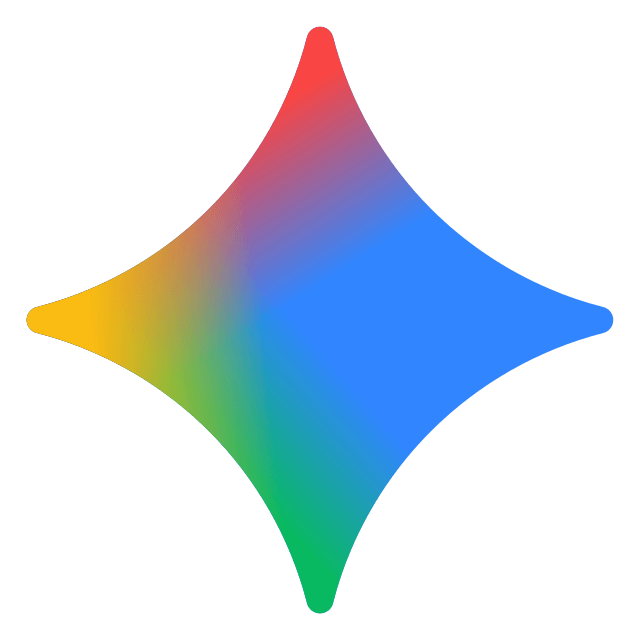}~Gemini-3-Pro & 55.2 & 47.8 & 32.1 & \textbf{71.3} & 55.0 & 54.0 & 44.8 & 57.1 & 79.6 \\
\midrule
% ======================== w./ World2Mind ========================
\rowcolor{opensrcHdr!75}
\multicolumn{10}{c}{\textbf{\textit{Frontier models w./ World2Mind}}} \\
% \modelicon{icons/qwen.png}~Qwen-3.5-397B & & &  &  &  &  &  &  & \\
\modelicon{icons/gpt.png}~GPT-5.2 & 54.0~{\scriptsize \textcolor{green!60!black}{($\uparrow$7.3)}} & 47.4~{\scriptsize \textcolor{red}{($\downarrow$5.1)}} & 33.4~{\scriptsize \textcolor{red}{($\downarrow$1.5)}} & 63.3~{\scriptsize \textcolor{red}{($\downarrow$4.2)}} & 52.4~{\scriptsize \textcolor{green!60!black}{($\uparrow$1.8)}} & \textbf{64.0}~{\scriptsize \textcolor{green!60!black}{($\uparrow$22.0)}} & 41.1~{\scriptsize \textcolor{green!60!black}{($\uparrow$0.4)}} & 51.0~{\scriptsize \textcolor{green!60!black}{($\uparrow$16.3)}} & 79.6~{\scriptsize \textcolor{green!60!black}{($\uparrow$28.6)}} \\
\modelicon{icons/claude.png}~Claude-4.6-Opus & 56.0~{\scriptsize \textcolor{green!60!black}{($\uparrow$17.7)}} & \textbf{59.0}~{\scriptsize \textcolor{green!60!black}{($\uparrow$12.0)}} & 34.3~{\scriptsize \textcolor{green!60!black}{($\uparrow$15.8)}} & 67.3~{\scriptsize \textcolor{green!60!black}{($\uparrow$5.2)}} & 54.8~{\scriptsize \textcolor{green!60!black}{($\uparrow$28.0)}} & \textbf{64.0}~{\scriptsize \textcolor{green!60!black}{($\uparrow$24.0)}} & 62.7~{\scriptsize \textcolor{green!60!black}{($\uparrow$15.6)}} & \textbf{65.3}~{\scriptsize \textcolor{green!60!black}{($\uparrow$30.6)}} & 40.8~{\scriptsize \textcolor{green!60!black}{($\uparrow$10.2)}} \\
\modelicon{icons/gemini.png}~Gemini-3-Pro & \textbf{61.0}~{\scriptsize \textcolor{green!60!black}{($\uparrow$5.8)}} & 51.8~{\scriptsize \textcolor{green!60!black}{($\uparrow$4.1)}} & \textbf{36.8}~{\scriptsize \textcolor{green!60!black}{($\uparrow$4.7)}} & 57.7~{\scriptsize \textcolor{red}{($\downarrow$13.5)}} & \textbf{62.6}~{\scriptsize \textcolor{green!60!black}{($\uparrow$7.6)}} & 62.0~{\scriptsize \textcolor{green!60!black}{($\uparrow$8.0)}} & \textbf{67.7}~{\scriptsize \textcolor{green!60!black}{($\uparrow$22.9)}} & \textbf{65.3}~{\scriptsize \textcolor{green!60!black}{($\uparrow$8.2)}} & \textbf{83.7}~{\scriptsize \textcolor{green!60!black}{($\uparrow$4.1)}} \\
\bottomrule
\end{tabular}
\vspace{-0.5cm}
\end{adjustbox}
\end{table*}

% To provide robust geometric-topological information, we formally define an \textbf{Allocentric-Spatial Tree (AST)} as the core spatial representation. The AST is a directed acyclic graph that utilizes geometrically stable landmarks (\eg, beds, tables) as core nodes to hierarchically associate surrounding smaller instances. Crucially, to approximate the fuzzy nature of human cognition, the AST employs rectangle-elliptical parameters (bounding box, major and minor axes, eccentricity, and rotation angle) to model the spatial footprints of instances. These designs equip foundation models with robust, dense, and highly actionable geometric-topological priors.

To provide robust geometric-topological priors, we formally define the \textbf{Allocentric-Spatial Tree (AST)} as the core spatial representation in World2Mind. The AST is a directed acyclic graph utilizing geometrically stable landmarks (\eg, beds, tables) as core nodes to hierarchically associate surrounding smaller instances. Crucially, to approximate the fuzzy nature of human cognition, the AST models spatial footprints using rectangle-elliptical parameters (bounding boxes, major/minor axes, eccentricity, and rotation angles). These designs equip models with robust, dense, and highly actionable geometric-topological priors.

However, merely offering spatial representation is insufficient to guarantee robust reasoning. In complex physical scenarios, reconstruction quality often suffers severe corruption due to occlusions or restricted viewpoints, leading to conflicts with objective geometric laws and raw visual observations. To mitigate this risk, we integrate a rigorous \textbf{\textit{spatial reasoning chain}} into World2Mind: \textbf{\textit{1)}} Difficulty Assessment and Tool Invocation, preventing over-computation on simple superficial queries; \textbf{\textit{2)}} Modality-Decoupled Cue Collection, independently extracting information from egocentric vision, AST structured text, and map visualizations; and \textbf{\textit{3)}} Geometry-Semantics Interwoven Reasoning, guiding the model to resolve cross-modal conflicts proactively and ultimately yield reliable spatial decisions.

Extensive evaluations across various spatial reasoning benchmarks demonstrate that World2Mind yields stable performance improvements of \textbf{6\%--18\%} for frontier models like GPT-5.2, while maintaining exceptional efficiency and reasoning interpretability. Astonishingly, leveraging the pure, high-density allocentric priors provided by the AST, text-only foundation models can execute complex 3D reasoning directly within their parameter space simply by reading the AST representation, approaching the performance of advanced multimodal models. Our findings offer a highly promising pathway to overcome the spatial cognition bottleneck in foundation models.

\section{Method Overview}
\label{sec:formatting}

\begin{table}[t]
\centering
\caption{\textbf{Results on the MindCube-Tiny~\cite{liu2024mindcube} benchmark.}}
\label{table:mindcube}
\renewcommand{\arraystretch}{1.0}
\setlength{\tabcolsep}{2pt}
\small
\begin{adjustbox}{width=\linewidth}
\begin{tabular}{@{}l cccc@{}}
\toprule
\textbf{Models} & \textbf{Avg.} & \textbf{Around} & \textbf{Among} & \textbf{Rotation} \\
\midrule
% ======================== Baseline ========================
\rowcolor{propriHdr!75}
\multicolumn{5}{c}{\textbf{\textit{Frontier models w/o. World2Mind}}} \\
\modelicon{icons/gpt.png}~GPT-5.2 & 49.9 & 62.4 & 45.2 & 48.5 \\
\modelicon{icons/claude.png}~Claude-4.6-Opus & 48.5 & 58.8 & 50.7 & 29.0 \\
\modelicon{icons/gemini.png}~Gemini-3-Pro & 75.1 & 77.2 & 68.2 & 93.0 \\
\midrule
% ======================== w./ World2Mind ========================
\rowcolor{opensrcHdr!75}
\multicolumn{5}{c}{\textbf{\textit{Frontier models w./ World2Mind}}} \\
\modelicon{icons/gpt.png}~GPT-5.2 & 54.6~{\scriptsize \textcolor{green!60!black}{($\uparrow$4.7)}} & 60.4~{\scriptsize \textcolor{red}{($\downarrow$2.0)}} & 47.7~{\scriptsize \textcolor{green!60!black}{($\uparrow$2.5)}} & 68.0~{\scriptsize \textcolor{green!60!black}{($\uparrow$19.5)}} \\
\modelicon{icons/claude.png}~Claude-4.6-Opus & 62.9~{\scriptsize \textcolor{green!60!black}{($\uparrow$14.4)}} & 82.4~{\scriptsize \textcolor{green!60!black}{($\uparrow$23.6)}} & 60.8~{\scriptsize \textcolor{green!60!black}{($\uparrow$10.1)}} & 45.0~{\scriptsize \textcolor{green!60!black}{($\uparrow$16.0)}} \\
\modelicon{icons/gemini.png}~Gemini-3-Pro & \textbf{81.6}~{\scriptsize \textcolor{green!60!black}{($\uparrow$6.5)}} & \textbf{86.0}~{\scriptsize \textcolor{green!60!black}{($\uparrow$8.8)}} & \textbf{75.8}~{\scriptsize \textcolor{green!60!black}{($\uparrow$7.6)}} & \textbf{93.5}~{\scriptsize \textcolor{green!60!black}{($\uparrow$0.5)}} \\
\bottomrule
\end{tabular}
\end{adjustbox}
\end{table}

This section details the technical overview of the proposed world2mind, as illustrated in Fig.~\ref{fig:teaser}

\subsection{Geometry-Semantic Alignment Pipeline}
Given an egocentric video sequence or multi-view image set $\{I_t\}_{t=1}^T$, our primary objective is to transcend the limitations of 2D vision and construct a robust 3D semantic representation of the physical world.

\noindent\textbf{\ding{182} Depth Estimation \& Semantic Extraction.} We employ \textbf{\textit{Depth Anything V3}}~\cite{lin2025depth} for monocular depth estimation, obtaining the depth map $D_t \in \mathbb{R}^{H \times W}$ and camera pose $T_t \in SE(3)$. Concurrently, we utilize \textbf{\textit{SAM3}}~\cite{carion2025sam} to extract open-vocabulary semantic masks $M_t$ based on a user-specified category list $\mathcal{C}$. To suppress the accumulation of long-tail errors inherent in depth estimation, we introduce a dual-level filtering mechanism based on the predicted confidence map $C_t \in [0,1]^{H \times W}$. Specifically, we formulate a binary validity mask $\mathcal{V}_t \in \{0,1\}^{H \times W}$ as follows:
\begin{equation}
\mathcal{V}_{t}(u, v)=\mathbb{I}\left(C_{t}(u, v)>\tau_{\text {pixel }}\right) \cdot \mathbb{I}\left(\mu_{t}>\tau_{\text {frame }}\right),
\end{equation}
where $\mu_t = \frac{1}{HW}\sum_{x=1}^{H}\sum_{y=1}^{W}C_t(x,y)$ denotes the global spatial confidence of frame $t$, and $\mathbb{I}(\cdot)$ is the indicator function that returns $1$ if the condition is met and $0$ otherwise. The variables $\tau_{\text{pixel}}$ and $\tau_{\text{frame}}$ represent the pixel-level and frame-level thresholds, respectively. A pixel is incorporated into the subsequent reconstruction only when $\mathcal{V}_t(u,v) = 1$.

\noindent\textbf{\ding{183} Point Cloud Mapping \& Density Filtering.} Qualifying 2D pixels are back-projected into the world coordinate system via the camera intrinsic matrix $K$, generating a global point cloud $\mathcal{P} = \{(\mathbf{p}_i, s_i, \mathbf{rgb}_i)\}_{i=1}^M$ carrying semantic labels $s_i \in \mathcal{C}$. Addressing the boundary outliers inherent in depth estimation, we propose a core region extraction strategy: for each point, we calculate its K-nearest neighbor local density $\rho_i = \frac{1}{K}\sum_{j \in \mathcal{N}_K(i)} \|\mathbf{p}_i - \mathbf{p}_j\|^{-1}$, and eliminate low-density "tail" points based on density percentiles. This yields an exceptionally pure geometry-semantic substrate.

\subsection{Allocentric Cognitive Mapping}

Inspired by the spatial mapping mechanisms of BI, we distill the unstructured point cloud into two highly abstract cognitive maps, enabling the model to proactively acquire spatial knowledge on demand via tool invocation.

\noindent\textbf{\ding{182} Landmark Cognitive Mapping.} Traditional methods rely on ambiguous relative relations or simplified grid representation~\cite{liu2024mindcube,yang2025thinking}. To overcome this, we formally define the \textbf{\textit{Allocentric-Spatial Tree (AST)}}, which reorganizes spatial entities as a directed acyclic graph within an absolute coordinate system. Specifically, we perform adaptive DBSCAN clustering on each semantic category within the point cloud to separate distinct instances. For each instance node, the AST discards traditional bounding boxes and instead fits a minimum bounding ellipse in the top-down view (X-Z plane), extracting the centroid $(x_c, z_c)$, major and minor axes $a$ and $b$, and rotation angle $\theta$. This parameterization: \textbf{\textit{1)}} significantly enhances robustness against reconstruction boundary noise; \textbf{\textit{2)}} perfectly aligns with the fuzzy probability nature of human spatial footprint perception. Output as dense structured text (e.g., YAML), the AST explicitly encodes hierarchical containment relationships among entities along with multi-dimensional geometric attributes.

\noindent\textbf{\ding{183} Route Cognitive Mapping.} For navigation-oriented tasks, World2Mind also enables extracting the masks of traversable categories (e.g., floors), back-projects and voxelizes them, and subsequently partitions them into an $N \times N$ grid map on the top-down plane. Combined with the mapping of the camera trajectory sequence $\{T_t\}$, this route map provides the model with explicit priors regarding passability and the human observer's motion trajectory.

\subsection{Geometry-Semantics Interwoven Reasoning}

In physical scenarios, 2D visual observations are susceptible to occlusions and adverse viewpoints, while 3D reconstruction information may contain local errors. To resolve potential contradictions between these two modalities, we design a rigorous three-stage interwoven reasoning chain.

\begin{figure}[t]
    \centering
    \includegraphics[width=0.44\textwidth]{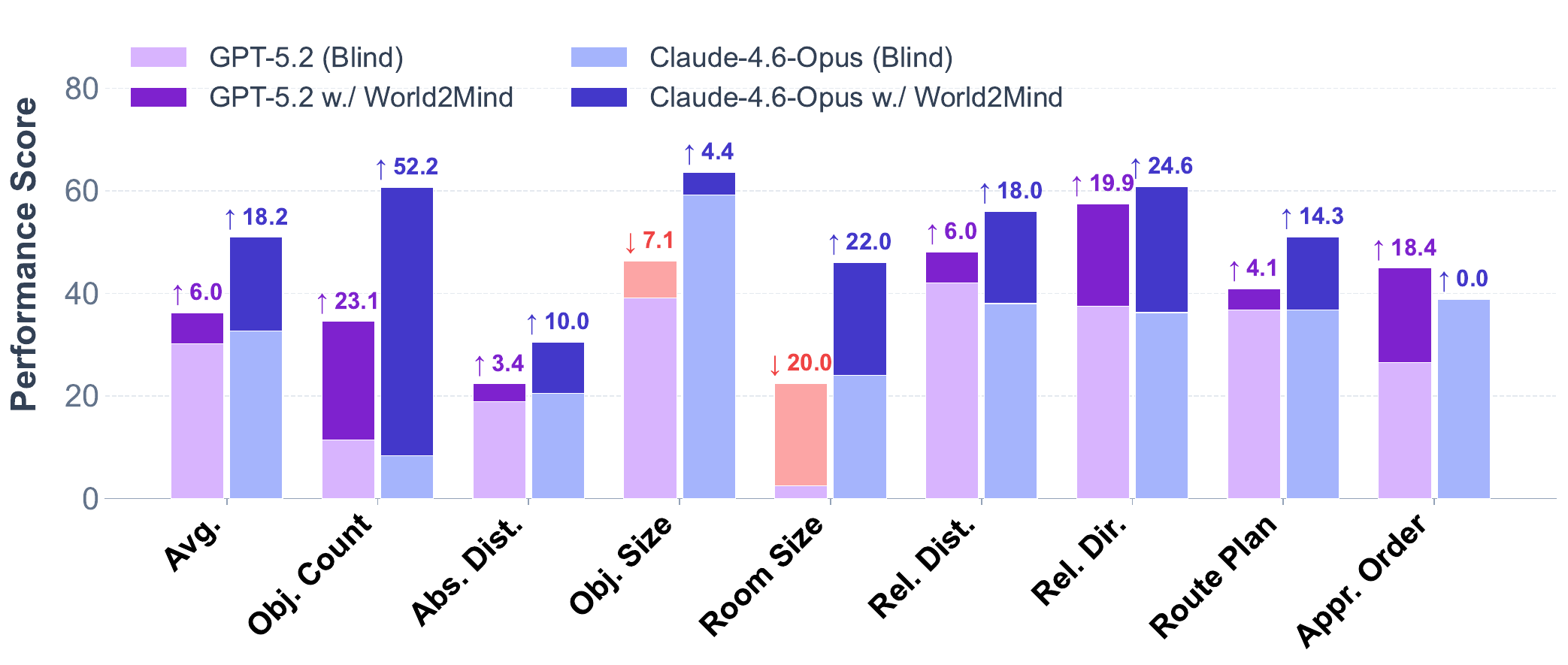}
    \caption{\textbf{Performance comparison under the text-only model (``blind") setting.} we report the performance gap on the VSI-Bench (Tiny subset) between foundation models relying solely on commonsense reasoning and those leveraging world2mind to acquire structured spatial knowledge for allocentric reasoning.}
    \label{fig:abla}
    \vspace{-0.4cm}
\end{figure}

\noindent\textbf{\textit{Stage 1: Tool Invocation Judgement.}} To reduce unnecessary computational overhead, the model must first evaluate the spatial relevance of the query. The model should proactively invoke World2Mind only when the task explicitly involves spatial reasoning, such as occlusion inference, distance estimation, or path planning.

\noindent\textbf{\textit{Stage 2: Modality-Decoupled Cue Collection.}} We force the model to extract information independently to prevent early modality bias. The model must simultaneously gather corroborating evidence from three independent sources: egocentric vision, the AST text returned by World2Mind, and optional 2D top-down map visualizations.

\noindent\textbf{\textit{Stage 3: Conflict Resolution and Cross-Validation.}} This is the crux of the reasoning chain. The model needs to proactively coordinate evidence across different modalities and identify cross-modal conflicts, such as missing objects due to visual truncation or coordinate drift caused by depth errors. By cross-validating the objective geometric parameters of the AST against subjective visual appearances, the model can dynamically weigh the credibility between visual illusions and reconstruction artifacts, ultimately outputting highly reliable and logically interpretable decisions.

\section{Experiment}

\subsection{Experimental Setup}

We conduct evaluations on two challenging spatial reasoning benchmarks: VSI-Bench~\cite{yang2025thinking}, which focuses on video-based reasoning in real-world physical scenes, and MindCube~\cite{liu2024mindcube}, which emphasizes multi-view cognitive mapping and mental simulation. Our evaluation primarily targets the most frontier multimodal foundation models, including GPT, Claude, and Gemini, given their exceptional proficiency in tool invocation and instruction-following.

\begin{figure}[t]
    \centering
    \includegraphics[width=0.46\textwidth]{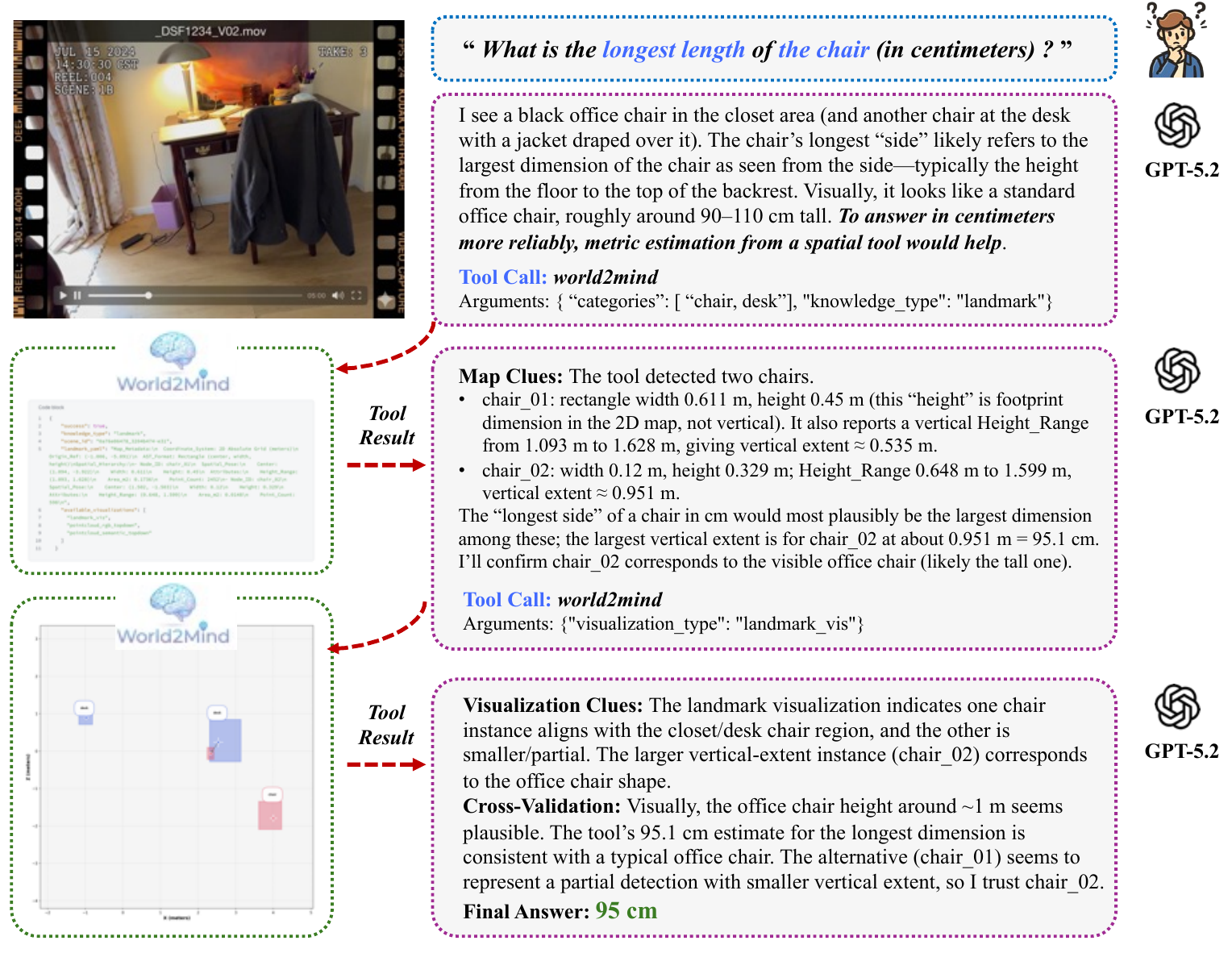}
    \caption{Complete reasoning trace under World2Mind.}
    \label{fig:case}
    \vspace{-0.4cm}
\end{figure}

\subsection{Main Results and Analysis}

\noindent\textbf{Evaluation on VSI-Bench.} As shown in Tab.~\ref{table:visbench_tiny}, the seamless integration of World2Mind yields a substantial leap in the average performance (Avg.) across all frontier models. Specifically, GPT-5.2 achieves a 7.3\% improvement, while Claude-4.6-Opus achieves a staggering 17.7\% improvement. In-depth analysis of the subtasks reveals that the performance gains are most striking in task categories that rely heavily on allocentric priors. For instance, on Relative Direction, Route Planning, and Relative Distance, the performance of Claude-4.6-Opus skyrockets by 15.6\%, 30.6\%, and 24.0\%, respectively. This compellingly demonstrates the critical role of the allocentric spatial knowledge provided by World2Mind in bridging the spatial reasoning gap.

\noindent\textbf{Evaluation on MindCube.} The results in Tab.~\ref{table:visbench_tiny} further corroborate the universality and robustness of our framework in sparse multi-view inputs. Even for Gemini-3-Pro, whose native spatial reasoning capability is already top-tier (with a baseline Avg. of 75.1\%), World2Mind successfully shatters its performance ceiling, pushing its average accuracy to 81.6\% (+6.5\%). Notably, in tasks like ``Rotation" that severely test 3D spatial imagination, the model achieves a remarkable performance breakthrough (GPT-5.2 improves by 19.5\%) due to its ability to perform logical deduction grounded in the AST.

\subsection{Ablation and Case Study}

To explore the limits of how structured text of AST empowers the spatial cognition of large language models, we follow \cite{yang2025thinking} to conduct ablation studies under the text-only (``blind") setting (see Fig.~\ref{fig:abla}). When visual image inputs are completely stripped away, foundation models that rely solely on commonsense priors degrade to near-random guessing on spatial tasks. Astonishingly, however, when equipped with World2Mind, both GPT-5.2 and Claude-4.6-Opus exhibit a remarkable performance rebound in the "blind" state. On core reasoning tasks such as Object Size and Route Planning, their scores closely approach those achieved with full visual inputs. This profound finding \textbf{\textit{indicates that pure, high-quality allocentric geometric priors are entirely sufficient to ignite powerful 3D mental reconstruction and simulation capabilities of foundation models under the text-based reasoning.}} Furthermore, we visualize the complete reasoning traces powered by World2Mind in Fig.~\ref{fig:case}, which clearly demonstrate that the interwoven reasoning process exhibits exceptional robustness and logical interpretability when resolving cross-modal conflicts.
{
    \small
    \bibliographystyle{ieeenat_fullname}
    \bibliography{main}
}


\end{document}